\documentclass[runningheads]{llncs}

\usepackage{graphicx}
\usepackage{multirow}
\usepackage[table,xcdraw]{xcolor}
\usepackage{amssymb}
\usepackage{subcaption}
\usepackage{float}
\usepackage[normalem]{ulem}
\useunder{\uline}{\ul}{}

\newcolumntype{L}[1]{>{\hsize=#1\hsize\raggedright\arraybackslash}X}

\usepackage{tabularx}
\usepackage{booktabs}
\usepackage[linkcolor=blue]{hyperref}
\usepackage{tikz}
\usepackage{cleveref}
\usepackage{paralist}
\usepackage{multirow}
\usepackage{paralist}
\usepackage{doi}
\usepackage[strings]{underscore}

\newcommand{\mytoprule}{\specialrule{\heavyrulewidth}{\aboverulesep}{0em}}
\newcommand{\mymidruletop}{\specialrule{\lightrulewidth}{\aboverulesep}{0em}}
\newcommand{\mymidrulebottom}{\specialrule{\lightrulewidth}{0em}{\belowrulesep}}
\newcommand{\myvertsp}{{\large$\phantom{\mid}$\hspace{-.25em}}}
\newcommand{\myhlemphrow}{\rowcolor{lightgray}\myvertsp}
\newcommand{\myhlrow}{\rowcolor[HTML]{EFEFEF}\myvertsp}
\newcommand{\myhlemphcell}{\cellcolor{lightgray}\myvertsp}

\begin{document}
\title{Challenges in the Safety-Security Co-Assurance of Collaborative Industrial Robots
}
\titlerunning{Safety-Security Co-Assurance for Collaborative Industrial Robots}
\author{Mario Gleirscher\inst{1}\orcidID{0000-0002-9445-6863}\and
Nikita Johnson\inst{1}\and
Panayiotis Karachristou\inst{2}\and
Radu Calinescu\inst{1}\and
James Law\inst{2}\and
John Clark\inst{2}}
\authorrunning{M. Gleirscher et al.}
\institute{Department of Computer Science, University of York, U.K. \\
\email{\{author.surname\}@york.ac.uk}
\and
Department of Computer Science, University of Sheffield, U.K. \\
\email{\{author.surname\}@sheffield.ac.uk}}

\maketitle              %

\begin{abstract}
The coordinated assurance of interrelated critical properties, such as
system safety and cyber-security, is one of the toughest challenges in
critical systems engineering.  In this chapter, we summarise
approaches to the coordinated assurance of safety and security.  Then,
we highlight the state of the art and recent challenges in human-robot
collaboration in manufacturing both from a safety and security
perspective.  We conclude with a list of procedural and technological
issues to be tackled in the coordinated assurance of collaborative
industrial robots.


 \end{abstract}

\section{Introduction}
\label{sec:introduction}

Collaborative robots~(or \emph{cobots}\footnote{Throughout this
  chapter we refer to \emph{cobots} as robots, including their software and
  operational infrastructure, specifically designed for
  human-robot collaboration, although we accept that traditional
  robots may be used in collaborative ways when augmented with
  sufficient control and sensory systems.}) are expected to drive the
robotics market in coming years~\cite{tantawi2019advances}, providing
affordable, flexible, and simple-to-integrate robotic solutions to
traditionally manual processes.  This transformational technology will
create new opportunities in existing markets such as food,
agriculture, construction, textiles, and craft
industries~\cite{hinojosa2018advanced,pawar2016manufacturing},
enabling more efficiency in production while reducing operator
workloads and removing occupational
hazards~\cite{hinojosa2018advanced,tang2019human}.

Cobots will ultimately enable humans and robots to share physical spaces, and combine the benefits of automated and manual processes~\cite{vanderborght2019unlocking}. However, current applications are in the main limited to those requiring little physical collaboration, with humans and robots sharing spaces but working sequentially~\cite{IFR2018demistifying};  close, physical collaboration in the true sense (with robots responding in real-time to users) requires more complex sensing and control, resulting in highly complex safety cases. Whilst cobots may be designed to be inherently safe (when operating with limited capabilities), the process (and end effector or payload) often poses a greater threat than the robot itself~\cite{tang2019human}.

ISO~10218~\cite{ISO10218} set the standard for safety of industrial
robots.  In 2016, with a rapidly growing range of collaborative robots
on the market, ISO~10218 was supplemented by ISO/TS 15066~\cite{ISO15066}, which
specifies additional safety requirements for industrial
\textit{collaborative} robots.  However, many manufacturing
organisations have indicated that these do not go far enough in
providing guidance, and that the lack of examples of good practice is
hindering deployment of the technology.  As a result companies are
falling back on traditional segregational approaches to risk
assurance, including physical or chronological isolation and barriers,
which counteract many of the lauded benefits of collaborative robots.

The aim of this chapter is to explore existing approaches and best-practice for safety and security of collaborative robots, and to highlight the challenges. \Cref{sec:generalApproaches} provides an overview of safety and security approaches applicable to cobots. Sections~\ref{sec:safety} and~\ref{sec:security} then elaborate on particular methodologies in the context of an industrial case study. Following the two perspectives---social and technical---of the Socio-Technical System~(STS) design approach~\cite{bostrom1977mis}, \Cref{sec:safesecure} enumerates additional socio-technical and technical challenges arising from safety-security interactions.

    \subsection{General Approaches to Safety and Security}
    \label{sec:generalApproaches}

One of the main sources of difficulty, which prompts the need for
guidance, is the engineering complexity and diversity required to
build a cobot, or indeed any complex system that involves
interactions with humans. In order to reconcile multiple (often
heterogeneous) goals and objectives, knowledge from multiple
engineering disciplines is needed -- mechanical, electrical, process,
human-computer interaction, and safety and security. In this section,
approaches to co-engineering and co-assuring safety and security that
can be applied to cobots are explored.  The discussion in this section will be
an extension of previous work done in~\cite{johnson2019assurance} with an emphasis on literature and application to industrial control, robotics and cobots. Throughout this section, \emph{assurance} will refer to the process and outcome of identifying, reducing and arguing about the level of risk.

Pietre-Cambacedes et al.~\cite{pietre2013cross} give a comprehensive
view of methods, models, tools and techniques that have initially been
created in either safety or security engineering and that have then
been transposed to the other quality attribute. However, one of the
biggest challenges identified is the (mis-)use of risk
language. Whilst both deal with the notion of risk, and aim to prevent
situations which might result in negative
consequences~\cite{eames1999integration}, their conditions and
concepts of loss are sufficiently different to cause conflict to
arise. A classical example of this is the conceptualisation of
risk. Traditionally, the goal of safety is to prevent death or injury,
therefore safety risk is concerned only with the hazards that might
lead to an accident, and is calculated as a product of the likelihood
of a hazard occurring and the severity of that hazard (i.e.~safety
risk = likelihood $\times$ severity). In contrast, the goal of
security is to prevent loss of assets. These might include people, in
which case the goal aligns with safety, however security encompasses a
much larger scope so assets might include process, intellectual
property, organisation reputation, and information. Thus many more
factors must be included when analysing security risk (i.e.~security
risk = (threat $\times$ access) $\times$ (business impact,
confidentiality, \textit{etc.})).

Figure \ref{fig:ssafpoints}, taken from previous work done on
co-assurance, shows a model from the Safety-Security Assurance
Framework (SSAF)~\cite{johnson2019assurance}. In particular, it
highlights the safety and security processes and interactions during
the system lifetime. Each of the phases of the system will place
different requirements on safety and security practitioners, for
example early stages will have a focus on assisting the
engineering process to lower the risk of the system, whereas during
operation, the practitioner's focus will be on ensuring that
operations are being performed as expected and no assurance claims
made earlier are being violated.

The core idea underlying SSAF is that of \textit{independent
  co-assurance} that allows for separate working but requires
\emph{synchronisation points} where information is exchanged and
trade-off decisions occur. This allows practitioners to use
specialised expertise and progress to occur within each domain because
there is a shared understanding of \textit{what} information will be
required, and \textit{where} and \textit{when} it should be provided.

There are different modes of interaction ranging from silos
(characterised by very few synchronisation points and little
inter-domain communication) to unified approaches (where the
attributes are co-engineered and co-assured together, e.g.\
in~\cite{Sabaliauskaite2015-AligningCyberPhysical}). A prerequisite
for knowing information needs at synchronisation points is
understanding the causal relationships within and between
domains. This can be achieved using multiple approaches, some of which
are shown in \Cref{tab:approachesTable} which contains a subset of
approaches that are commonly used.  Approaches to safety
and security risk management can be classified into several groups
according to their level of formalism and their objectives. The
classification framework of \textit{(1)} structured risk analysis,
\textit{(2)} architectural methods, and \textit{(3)} assurance and
standards has been adapted from~\cite{pietre2013cross}:

\begin{figure}
\centering
\includegraphics[width=0.8\textwidth]{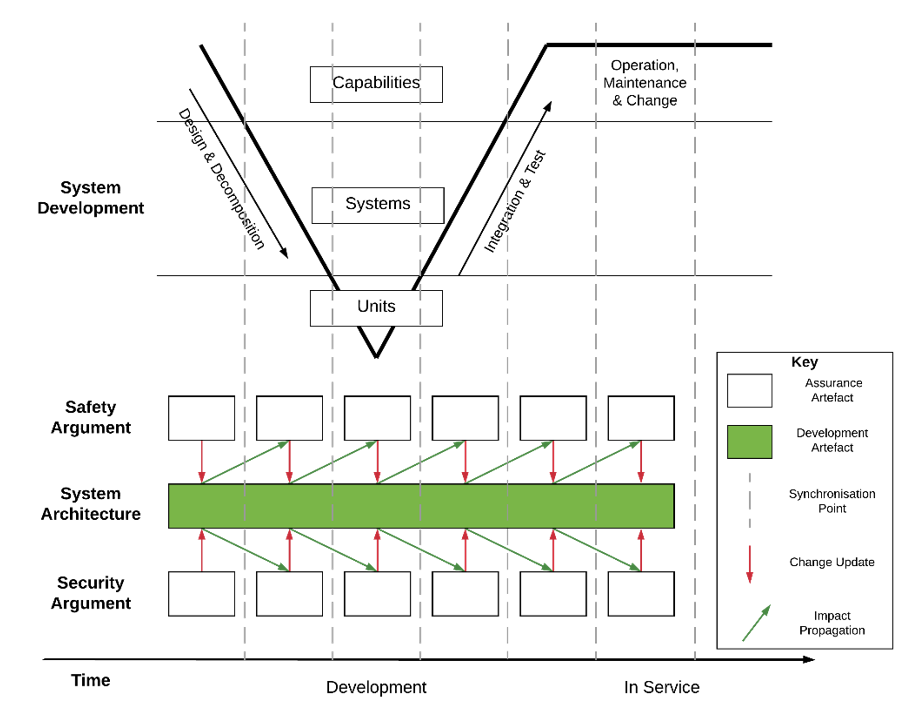}
\caption{SSAF Interaction Model with Synchronisation Points (taken from \cite{johnson2019assurance}).}
\label{fig:ssafpoints}
\end{figure}

\begin{table}
\centering
\caption{Approaches to Safety and Security Assurance Applicable to Cobots.}
\label{tab:approachesTable}
\begin{tabularx}{\textwidth}{%
    >{\bfseries\hsize=.4\hsize}X
    >{\hsize=.2\hsize}X
    >{\hsize=.2\hsize}X
    >{\hsize=.2\hsize}X}
\mytoprule
\textbf{Approach} & \textbf{Safety} & \textbf{Security} & \textbf{Joint} \\ \mymidruletop
\multicolumn{4}{l}{\myhlemphcell\textit{(1) Structured Risk Analysis}}                    \\ 
\myhlrow
HAZOP     &     \cite{dunjo2010hazard}  &   \cite{srivatanakul2004effective}   &    \cite{winther2001security}     \\ 
FME(C)A  &    \cite{gilchrist1993modelling} \cite{bouti1994state}             &                   & \cite{schmittner2014fmvea} \\ 
\myhlrow 
Bowtie Analysis  &  \cite{yu2017bow} & \cite{cook2016measuring}  &      \cite{abdo2018safety} \cite{despotou2009addressing}            \\ 
STPA  & \cite{leveson2003new}    &                   &    \cite{young2014integrated} \cite{friedberg2017stpa}          \\ 
\myhlrow 
STRIDE  &    &      \cite{martin2018quantitative}    &                  \\ 
FTA &   \cite{handbook1981nureg}     &      &
\cite{johnson2019through} \cite{Sabaliauskaite2015-AligningCyberPhysical} \\ 
\myhlrow 
Attack Trees &     &      \cite{schneier1999attack} \cite{kordy2010foundations}            &  \cite{Sabaliauskaite2015-AligningCyberPhysical}                \\  
\multicolumn{4}{l}{\myhlemphcell\textit{(2) Architectural Approaches, Testing \& Monitoring}}                      \\ 
\myhlrow 
ATAM  &                 &                   &    \cite{kazman1998architecture}             \\  
Metrics &                 &                   & \cite{knowles2015survey}     \\ 
\myhlrow 
Failure Injection & \cite{papadopoulos1999hierarchically}  & \cite{arkin2005software} &                  \\ 
\multicolumn{4}{l}{\myhlemphcell\textit{(3) Assurance \& Standards }}                         \\ 
\myhlrow 
Assurance Cases &   \cite{kelly1999arguing}         &       \cite{he2012generic}    &                  \\ 
 Standards &       IEC 61508        &     IEC 62443       &     IEC TR 63069             \\ 
  &    ISO 10218       &      ISO 15408 (CC)       &                \\
  & ISO/TS 15066   &      ISO 27001     &                \\ \bottomrule
\end{tabularx}
\end{table}

\subsubsection*{(1) Structured Risk Analysis.} 
\label{sec:str-rsk-ana}

These approaches are
concerned primarily with understanding the cause-effect relationships
for particular risks. In safety, hazards are analysed by experts using
structured reasoning approaches such as Hazard and Operability
Studies~(HAZOP), Failure Modes Effects and Criticality Analysis~(FMECA), and
Bow Tie Analysis. These function by requiring the analyst to consider
risk sources and outcomes systematically by using guide words over
system functions.  Example guidewords are \textit{too much},
\textit{too little}, \textit{too late}, \textit{etc.}; when applied to
a cobot speed function would allow the analyst to reason about what
would happen if the cobot was too fast/slow/late performing a task in
a specific context. System-Theoretic Process Analysis~(STPA) requires
the system to be modelled as a controlled process rather than decomposing it into functions. Hazards relating to the control structure are then reasoned about. The advantage of this approach is that hazards introduced by failure of intent are more readily identified than analysing functions, for example, where a cobot function does not fail and system requirements are satisfied, but it still leads to an unsafe state. There have been multiple adaptations of safety approaches to include security guidewords and prompts such as security-informed HAZOPs, FMVEA (FMEA and Vulnerabilities Analysis) and STPA-Sec. %

Similar to its safety counterparts, STRIDE is a security threat analysis model that assists analysts to reason about types of threats relating to spoofing, tampering, repudiation, information disclosure, denial of service and elevation of privilege (more information is provided in \Cref{sec:security}).  The main limitation of structured risk analysis is their over-reliance on expert opinion which makes the outcome only as good as the analyst performing the analysis~\cite{johnson2019assurance}. 

Fault Tree Analysis (FTA)~\cite{handbook1981nureg} is one of the
earliest approaches used in safety analysis. Its process requires that
the causes of functional failures and faults are decomposed; the
outcome is then modelled in a directed acyclic graph as events
connected by AND and OR gates. This tree-based approach has been
extended to apply to both attacks and threats in
security~\cite{schneier1999attack,kordy2010foundations}. Whilst the
formality of these types of models allows for an improved review and audit of the risk analysis, they are limited by the assumptions made during the analysis process and whether or not they are validated subsequent to the analysis.

\subsubsection*{(2) Architectural Approaches, Testing \& Monitoring.} 
\label{sec:arch-test-mon}

The third class of approaches seeks to understand the trade-offs between safety and security for the system architecture, and the effects of risk during operation. Architectural Trade-Off Analysis Method (ATAM)~\cite{kazman1998architecture} is a methodology that allows structured negotiation of architectural strategies for quality attributes including safety and security. Dependability Deviation Analysis~(DDA)~\cite{despotou2009addressing} relies on applying Bow-Tie analysis to architectural models to understand the impact of dependability attributes on each other. In addition to these approaches, which are primarily applied early in the system development lifecycle, there are approaches that seek to test and verify the requirements output at even earlier stages. In addition to architectural model approaches such as ATAM and DDA, the concept of joint safety and security metrics~\cite{knowles2015survey} has been proposed to understanding safety and security functional assurance in industrial control systems. Examples of failure injection and testing include  Hierarchically Performed Hazard Origin and Propagation Studies~(HiP-HOPS)~\cite{papadopoulos1999hierarchically} which is a safety technique wherein Simulink models are used to generate fault trees to assess the impact of different faults. In security, penetration testing~\cite{arkin2005software} is often used as an approach to understand the vulnerabilities in the engineered system.

\subsubsection*{(3) Assurance \& Standards.}
\label{sec:ass-std}

The final class of approaches seeks to manage risk through safety and
security processes and standards. The industrial control safety
standard IEC 61808 is arguably the most influential safety standard
created as many safety standards from other domains are its
derivatives; examples include ISO~10218~\cite{ISO10218}~(robotics) and
ISO/TS~15066~\cite{ISO15066}~(cobots). Similarly for security IEC 62443~(ICS product security), ISO 27001 (organisation security process), ISO 15408~(security requirements) have been applied to a wide range of industries. Standards from both domains have been largely isolated with very little alignment,  however, as security has the potential to undermine the safety of a system, the two should be considered in conjunction. In an attempt to address this problem, joint standards are being created. One such standard is IEC TR 63069~(applied to ICS), which advocates the creation of a \textit{security perimeter} within which safety analysis is performed. A fundamental part of the certification process for standards is presenting a structured argument with evidence to show that the assurance criteria have been satisfied. This argument takes the form of safety cases and security cases (examples of each in \cite{kelly1999arguing} and \cite{he2012generic}).

Many of the approaches described above are general approaches or have
been applied within an industry related to cobots, such as
industrial control. There has been some preliminary application of
many of these methods to cobots, for example, Lichte and
Wolf~\cite{lichte2018use} use an approach that combines graphical
formalism and architectural methods to understand the safety and
security of a cobot; this is a good start, however it still
leaves many challenges unaddressed. In the following sections, we
provide an illustrative cobot example and further details about the safety and security challenges.

\subsection{Illustrative Example: The Cobot}

In the previous section, the focus was on general approaches to safety and security that could be applied to a collaborative robotic system. However, there is a paucity of information about how these work in practice. The industrial case study discussed in the following Sections \ref{sec:safety} and \ref{sec:security}, seek to provide greater clarity about how to undertake safety and security analysis for a cobot.

As many types of human-robot interactions, Kaiser et al. describe \emph{scenarios of co-existence}~\cite{Kaiser2018-SafetyRelatedRisks}:
\begin{inparaenum}[(1)]
\item \emph{Encapsulation}, with physically fenced robot work areas,
\item \emph{co-existence}, without fencing but separation of human and
  robot work areas,
\item \emph{cooperation}, with shared work areas but without
  simultaneous operation in the shared area, and
 \item \emph{collaboration}, with shared work areas and simultaneous,
   potentially very close interaction inside the shared area.
\end{inparaenum}

In our example, we consider scenarios that match (3) or (4) and a
system that comprises primarily of a plant (i.e.~a manufacturing cell
including one or more robots), operators, and an automatic controller.

\section{Cobot Safety}\label{sec:cobotsafety}
\label{sec:safety}
In this section, we (i)~review safety risk analysis and handling in
human-robot collaboration and (ii)~highlight recent challenges in the
development of controllers implementing the safety requirements.

\subsection{Analysing Safety Risks in Cobot Settings}
\label{sec:revi-hazardsr-from}

In safety analysis, one focuses on two phenomena: \emph{accidents}
and \emph{hazards}.  An \emph{accident} is a more or less immediate
and undesired \emph{consequence} of a particular combination of the
plant's state and the environment's state, which together form the \emph{cause} of
the accident.  The plant's portion of such a cause is called a
\emph{hazard} and the environment's portion an \emph{environmental condition}~\cite[p.~184]{Leveson2012-EngineeringSaferWorld}, both first-class entities of safety risk analysis.

\subsubsection*{Safety Risks.}

Accidents, their causes, and hazards in human-robot co-existence in
manufacturing have been discussed in the literature since the mid
1970s (e.g.~\cite{Sugimoto1977-SafetyEngineeringIndustrial,Jones1986-StudySafetyProduction,Alami2006-Safedependablephysical,Kaiser2018-SafetyRelatedRisks}).
Notably, Nicolaisen~\cite{Nicolaisen1985-OccupationalSafetyIndustrial}
envisioned collaborative, potentially mobile, robots closely working
together with humans in shared work areas in 1985.
In 1986, Jones~\cite{Jones1986-StudySafetyProduction} distinguished two major
classes of hazards relevant to human-robot collaboration:
\begin{itemize}
\item \emph{Impact hazards:} Unexpected \emph{movements} (e.g.\
  effectors or joints reaching beyond the planned work area), for
  example due to failing equipment (e.g.\ valves, cables,
  electronics, programs); dangerous work pieces handled (e.g.\
  released) in an unexpected or erroneous manner; manipulations (e.g.\
  welding, cutting, transport) with hazardous side effects (e.g.\
  welding sparks, flying tools or work piece residuals, robots bumping
  into humans).
\item \emph{Trapping hazards:} Workers are trapped between a robot and a static object~(e.g. a machine or cage wall) while the robot is active or
  in a programming, maintenance, or teaching mode.
\end{itemize}
Both classes of hazards can cause the robot to collide with, or catch,
a human co-worker.  This includes robot joints, end-effectors, or dangerous work
pieces hitting and injuring co-workers.

\subsection{Handling Safety Risks in Cobot Settings}
\label{sec:revi-exist-cobot}

Accidents or hazards can be \emph{prevented} by employing
measures to avoid their occurrence.  More generally, accidents or hazards can be \emph{mitigated} by employing measures to
significantly reduce accident or hazard \emph{probability} or 
reduce accident \emph{severity}.  An accident cause is considered
\emph{latent}\footnote{As opposed to \emph{immediate} causes falling
  outside the scope of risk handling.} if there are sufficient
resources~(e.g. time, bespoke safety measures) to mitigate the accident~(e.g. by
removing the hazard or the environmental condition).  

For example, a robot arm on its way to a shared area~$W$~(hazard) and the operator on their way to~$W$~(environmental condition) form a \emph{cause} of an impact \emph{accident} to happen.  This cause seems \emph{immediate} unless a measure~(e.g.~stopping the arm on a collision signal, or the operator jumping away) renders it to be \emph{latent}, i.e.~possible if this measure were to stay inactive.

Measures can be either \emph{intrinsic}~(not requiring control equipment) or
\emph{functional}\footnote{Note that \emph{functional safety} in
  IEC~61508 or ISO~26262 deals with the dependability, particularly,
  correctness and reliability, of \emph{critical} programmable
  electronic systems.  Safety functions or, here, ``functional
  measures'' form the archetype of such systems.} (requiring control equipment).  A functional measure is said to be
\emph{passive} if it mitigates certain accidents~(e.g. a car airbag
system) and \emph{active} if it prevents certain accidents~(e.g. an
emergency braking system).  Functional measures focusing on the
correctness, reliability, or fault-tolerance of the controller~(nowadays a complex programmable electronic system) are called
\emph{dependability} measures~\cite{Alami2006-Safedependablephysical}.

\subsubsection*{Safety Measures.}

Experience from accidents and risk analyses has lead to many
technological
developments~\cite{Alami2006-Safedependablephysical,Hayes2013-ChallengesSharedEnvironment,Kaiser2018-SafetyRelatedRisks},
partly inspired by precautions when interacting with machinery.
\Cref{tab:measures} gives an overview of measures in use.

\begin{table}[t]
  \centering
    \caption{Examples of safety measures in human-robot
    collaboration, classified by stage of escalation and by
    the nature of the underlying technical mechanism}
  \label{tab:measures}
  \footnotesize
  \begin{tabularx}{\textwidth}{>{\bfseries}L{.15}L{.3}L{.2}L{.35}}
    \mytoprule
    \myhlemphrow
    \textbf{Stage}
    & \textbf{Type}
    & \textbf{Intrinsic}
    & \textbf{Functional}
    \\\mymidrulebottom
    Hazard prevention
    & 1) Safety barrier, physical safeguard
    & Fence
    & Interlock
    \\
    & 2) Avoidance of development mistakes
    & Controller verification
    & Controller verification
    \\\midrule
    \multirow{2}{2cm}{Hazard mitigation \&
    accident prevention}
    & 3) Improved reliability and
      fault-tolerance
    & 
    & Fault-tolerant scene interpretation 
    \\
    & 4) Intrusion detection (static, dynamic,
      distance-based)
    &
    & Speed \& separation monitoring;
      safety-rated monitored stop
    \\
    & 5) Hand-guided operation
    \\\midrule
    Accident mitigation
    & 6) Power and force limiting 
    & Lightweight components, flexible surfaces
    & Variable impedance control, touch-sensitive and force-feedback
      control 
    \\
    & 7) System halt
    &
    & Emergency stop
    \\\bottomrule
  \end{tabularx}
\end{table}

\subsubsection*{Concepts for Functional Measures.}

Jones~\cite[p.~100]{Jones1986-StudySafetyProduction} describes safety
monitors as an additional control device for safety actions like
emergency stop and limiters~(e.g. speed, force).
He further suggests safety measures be designed specifically for
each \emph{mode of operation} (e.g. programming or teaching,
normal working, and maintenance), each to be examined for its designed
and aberrant behaviour~\cite[p.~87]{Jones1986-StudySafetyProduction}.
Internal malfunction diagnostics~(e.g. programming error detection, electronic
fault detection, material wear-out monitoring) 
can also inform such a safety monitor and trigger
mode-specific actions.

Alami et al.~\cite{Alami2006-Safedependablephysical} highlight 
advantages of \emph{interaction-controlled} robots over 
\emph{posi\-tion-controlled} robots.  The former are easier for
planning and fewer assumptions on the structure of the work area and
the robot's actions need to be made, while the latter need extensive
pre-planning and require stronger assumptions.

Based on ISO~15066, Kaiser et
al.~\cite{Kaiser2018-SafetyRelatedRisks} summarise design
considerations~(e.g.\ work area layouts) and collaborative
\emph{operation modes} beyond traditionally required emergency stop
buttons~(also called dead man's controls):
\begin{itemize}
\item \emph{Safety-rated monitored stop}: A stop is assured while
  powered, thus prohibiting simultaneous operation of robot and
  operator in the shared area.
\item \emph{Hand-guiding operation}: The robot only exercises
  zero-gravity control and is guided by the operator, hence, no
  actuation without operator input.
\item \emph{Speed and separation monitoring}: Robot speed is
  continuously adapted based on several regions of distance between
  robot and operator.
\item \emph{Power and force limiting}: To reduce impact force on the
  human body, the robot's power and applied forces are limited.
\end{itemize}

\subsubsection*{Guidelines.}

Standardisation of safety requirements for industrial robots 
started in Japan with the work of
Sugimoto~\cite{Sugimoto1977-SafetyEngineeringIndustrial}.  In the meantime,
ANSI/RIA R15.06, ISO~10218, 13482, and 15066 have emerged, defining
safety requirements and providing guidance for safety measures.  The
aforementioned safety modes are part of the recommendations in
ISO~15066.

\subsection{Risk Analysis and Handling in a Cobot Example}
\label{sec:appl-cobots-proj}

In our example, an operator and a cobot undertake a spot-welding process.  The open-sided work cell consists of a robot positioned between a spot-welding machine and a shared hand-over area. During normal operation an operator sits outside the cell and uses this area for exchanging work pieces.  The layout allows staff to enter the active cell when needed.

In \Cref{tab:riskana}, we exemplify results of a preliminary safety
analysis~(i.e.\ hazard identification, assessment, requirements
derivation, cf.\ \Cref{sec:revi-hazardsr-from}) of the cell along with the identified safety requirements
compliant with state-of-the-art measures as summarised in
\Cref{sec:revi-exist-cobot}.  The right column specifies the
\emph{overall safety requirements} for each \emph{accident} and the
technical safety requirements~(e.g.\ the mode-switch requirements) for
each \emph{latent cause}~(left column) indicating how the
\emph{hazard} can be removed from the critical state in due time.
Each safety requirement specifies the conditions of a corresponding
measure to mitigate any escalation towards one of the listed
accidents.

\begin{table}[t]
  \centering
    \caption{Results of hazard identification, assessment, and
    mitigation analysis referring to measures recommended in ISO~15066}
  \label{tab:riskana}
  \footnotesize
  \begin{tabularx}{\textwidth}{>{\hsize=.35\hsize}X>{\hsize=.65\hsize}X}
    \toprule
    \textbf{Critical Event}
    &
    \textbf{Safety Requirement}$^{\dagger}$
    \\\hline
    \multicolumn{2}{l}{\myhlemphcell \textbf{Accident} (undesired, to be \emph{mitigated})}
    \\\hline
    The robot arm collides with the operator.
    & The robot shall \emph{avoid} active collisions with the
    operator.
    \\
    \rowcolor[HTML]{EFEFEF} 
    Welding sparks cause operator injuries
    (i.e. burns).
    & The welding machine \emph{avoids} sparks injuring the operator.
    \\\hline
    \multicolumn{2}{l}{\myhlemphcell \textbf{Latent Cause} (to be \emph{reacted
    upon} in a timely manner)}
    \\\hline
    The operator and the robot use the shared hand-over area at the same time.
    & (m) The robot shall perform a \emph{safety-rated monitored stop}
    and (r) resume \emph{normal operation} after the \emph{operator} has left the
    \emph{shared hand-over area}. 
    \\
    \rowcolor[HTML]{EFEFEF} 
    The operator approaches the shared hand-over area 
    while the robot is away from the hand-over area (undertaking a different part of the process).
    & (m) If the robot is transferring a work piece to the hand-over area then
    it shall switch to \emph{power and force limiting} mode and (r) resume
    \emph{normal operation} after the \emph{operator} has left the
    \emph{shared hand-over area}.
    \\
    The operator has entered the safeguarded area of the
    cell while the robot is moving or the welding process is active.
    & (m) The \emph{welding machine}, if running, shall be \emph{switched off} and the
    \emph{robot} shall switch to \emph{speed and separation
      monitoring} mode. (r) Both robot and welding machine shall resume normal
    operation after the operator has left the area and acknowledged the
    safety notification provided via the operator interface.
    \\
    \rowcolor[HTML]{EFEFEF} 
    The operator is close to the welding spot while the robot is
    working and the welding process is active.
    & (m) The \emph{welding machine} shall be \emph{switched off} and the \emph{robot}
    shall perform a \emph{safety-rated monitored stop}. (r) Both robot
    and welding machine shall resume \emph{normal
    operation or idle mode with a reset procedure} after the operator
      has left the area and acknowledged the 
    safety notification provided via the operator interface.
    \\\bottomrule
    \multicolumn{2}{p{.95\columnwidth}}{
      $^{\dagger}$
      (m) \dots mitigation,
      (r) \dots resumption
    }
  \end{tabularx}
\end{table}

\subsection{Recent Challenges in Cobot Safety}
\label{sec:chall-cobot-safety}

Because robots and human co-workers have complementary skills, it has
long been an \emph{unfulfilled desire for both actors to simultaneously and closely
work in shared areas}~\cite[p.~70]{Jones1986-StudySafetyProduction}.
Consequently, complex guarding arrangements, established for safety reasons, interfere 
with efficient production workflows.  For mobile robots, fencing in the
traditional sense is rarely an option.
Back in the 1980s, another problem was that robots often had to be
programmed through a teaching-by-demonstration approach with workers in the cage while robots 
fully powered.  However, this rather unsafe method has been superseded by more sophisticated
programming techniques~(e.g.~simulation, digital twins).

From the viewpoint of co-workers, close interaction has proven to be
difficult in some settings as invisible joint motion planning
procedures not rarely result in \emph{unpredictable joint/effector
  movement} patterns.  Assuming collaborative scenarios with complex
tasks shared between humans and robots, Hayes and
Scassellati~\cite{Hayes2013-ChallengesSharedEnvironment} raise further
conceptual issues, such as how cobots recognise operator intent, how
they autonomously take roles in a task, how they include operators'
intent in their trade-offs, and how they self-evaluate quantities such
as risk during operation.

Unfortunately, more sophisticated robots incorporate a larger variety
of more \emph{complex failure modes}.
Sensors need to sufficiently inform the robot controller for the
estimation and \emph{timely reduction of the impact on the human
  body}.  For example, ``speed and separation monitoring''
requires \emph{accurate and reliable sensors}~(e.g.\ stereo-vision
systems, laser scanners), particularly, when multiple safety zones are
considered, to be able to control several modes of operation.
Unknown faults, complex failure modes, and additional sensor inputs lead to \emph{security} as a more than ever important prerequisite to assuring cobot safety, as pointed out by Kaiser et al.~\cite{Kaiser2018-SafetyRelatedRisks}.
Finally, the overarching challenge is to \emph{provide practical safety and performance
  guarantees of the behaviour of a manufacturing cell} based on
realistic assumptions and leading to certifiable assurance cases.

\section{Security} \label{sec:security}
\label{sec:security}
As indicated in \Cref{sec:cobotsafety}, an important aim of 
research has been to facilitate the release of cobots from highly
constrained caged environments to enable greater
productivity~\cite{matheson2019human}. Ensuring their security is a
critical aspect. The security of robots has received less attention
than safety concerns and cobot-specific security issues have received
hardly any attention at all.  We can learn from extant attempts to
secure robots in general~(though this area itself is still a
challenge) and seek to identify new challenges for cobots.
 
Some security issues have direct safety implications, for example any
attack that can gain control over the physical actions of a cobot may
cause the robot to behave in a physically dangerous fashion, even to
the point of launching a physical attack on the co-worker. Some, for
example the leakage of personal data, may not have any obvious safety
effect, but must be addressed in a manner that satisfies security
related regulations~(e.g.\ GDPR). 
Below we examine elements of security whose challenges have particular
relevance to cobot security.

\subsection{Threat analysis}

\subsubsection*{Threat Modelling} is the use of abstractions which aid
the process of finding security risks. The result of this operation is
often referred to as a threat model. In the case of robotics, threat
modelling defines risks that relate to the robot and its software and
hardware components while offering means to resolve or mitigate them
\cite{vilches_2019}.\footnote{We would add that analysing the wider
  system in which a cobot exists is essential, since for example
  linking OT and IT---e.g.\ supporting accounting and other business
  functions---is a common practice and greatly enlarges the threat
  landscape.} In general terms, threat modelling is figuring out what
might go wrong in terms of security with the system being built, and
helps address classes of attacks, resulting in the delivery of a much
more secure product \cite{shostack2014threat}.

An example of threat modelling used in various systems is STRIDE. Kohnfekder and Garg \cite{kohnfelder1999threats} introduced the STRIDE threat modelling approach in 1999  and it has since become one of the most commonly used threat modelling methods. The STRIDE acronym stands for Spoofing, Tampering,  Repudiation,  Denial of Service and Elevation of Privilege. This methodology has recently been used in a document concerning the threat modelling of robots \cite{ros2_Robotic_systems_threat_model} (at the time of writing this is still a draft document).

Moreover, a novel way of identifying attacks/threat modelling is used in a paper by Trend Micro \cite{maggi2017rogue}. They identify five robot specific attack classes and their concrete effects, as depicted in \Cref{ROUGETABLE}.

\begin{table}
\caption{Types of attacks to robots by Trend Micro
\cite{maggi2017rogue}}
\label{ROUGETABLE}
  \footnotesize
\begin{tabularx}{\textwidth}{XcccX}
  \mytoprule \myhlemphrow
  & \multicolumn{3}{c}{\cellcolor[HTML]{C0C0C0}{\ul\textbf{Requirements Violated}}}
  & \\
  \myhlemphrow
  \multirow{-2}{*}{\cellcolor[HTML]{C0C0C0}\textbf{Robot Attack
      Class}}
  & \cellcolor[HTML]{C0C0C0}\textbf{Safety}
  & \cellcolor[HTML]{C0C0C0}\textbf{Integrity}
  & \cellcolor[HTML]{C0C0C0}\textbf{Accuracy}
  & \multirow{-2}{*}{\cellcolor[HTML]{C0C0C0}\textbf{Concrete Effects}}
  \\ \mymidrulebottom
  1. Altering the control loop parameters
  & \checkmark & \checkmark
  & \checkmark
  & Products are modified or become defective          \\
  \rowcolor[HTML]{EFEFEF}
  2. Tampering with calibration parameters  & \checkmark                                                  & \checkmark                                                     & \checkmark                                                    & \cellcolor[HTML]{EFEFEF}Damage to robot                                                      \\
  3. Tampering with the production logic    & \checkmark                                                  & \checkmark                                                     & \checkmark                                                    & Products are modified or become defective          \\
  \rowcolor[HTML]{EFEFEF} 4. Altering the user perceived robot state &
  \checkmark & X & X
  & \cellcolor[HTML]{EFEFEF} Injuries to robot operator \\
  5. Altering the robot state & \checkmark & X & X & Injuries to robot
  operator \\ \bottomrule
\end{tabularx}
\end{table}

\subsection{Review of existing security approaches}

\subsubsection{Security Policies.}

For a system to be \emph{secure} there needs to be an unequivocal statement of what it means to be secure, what properties the system must uphold, what must happen and what must not happen. Such a statement is typically provided by a \emph{security policy}. The policy may cover physical (e.g. access to a physical robot environment may be required to be via specific doors accessible only to authorised users), logical (e.g. access control policy), and procedural aspects (e.g. vetting of staff). For example, Zong, Guo and Chen~\cite{zong2019policy} propose a policy-based access control system that enables permission management for robot applications developed with ROS (Robot Operating System). They introduce Android-like permissions so that each application in ROS can be controlled as to whether a certain resource can be accessed or performed.

\subsubsection{Authentication}
underpins security in various forms. It can be thought of as the
verification of one or more claims made about system agents or
elements to be valid, genuine or true. Most commonly, authentication applies to a user or a process that wishes to gain access to a system and its resources. Authentication may be required from the user to the system~(\emph{user
  identification}) or from the system to the user.\footnote{Though in
  many applications guarantees of authenticity of the system will be
  provided by physical and procedural means---for example, it may be
  practically very hard to replace an authentic robot with a malicious
  or fake one.}
Dieber et al.~\cite{dieber2017security} introduces a dedicated Authentication Server (AS) that tracks which ROS node subscribes or publishes in a topic. The aim of ROS is to provide a standard for the creation of robotic software that can be used on any robot. A node is a sub-part of this robotic software. The software contains a number of nodes that are placed in packages, two example nodes are the camera driver and the image processing software.  These nodes need to communicate with each other \cite{quigley2009ros}. The proposed AS also handles node authentication and produces topic-specific encryption keys.  

User authentication is typically provided in three ways:
\begin{inparaenum}[(i)]
\item \emph{something you know}, 
\item \emph{something you have}, and
\item \emph{something/someone you are}.
\end{inparaenum} Passwords are the most common means of user authentication. A
\emph{claim} is first made by supplying a specific user identifier, or
\emph{user\_id}~(the person supplying that user\_id is claiming to be
the person associated with it by the system). Accompanying such a
claim with a password recognised by the system is an example of (i).
(ii) is also a major means of authentication. Possession of a `token'~(or physical identifier) is deemed to establish the link with an authorised user. Thus, Radio
Frequency Identification Tags~(RFIDs) may be programmed with the
identities of specific users to whom they are given.
Biometric approaches are examples of~(iii). These make use of physical~(biological) or behavioural features of a user. Fingerprints,
voiceprints, retinal or iris images are all examples of biometrics
that can be used to identify a individual.

We would observe that most user authentication is \emph{one off}: the user authenticates at the beginning of a session and the privileges which go with such authentication prevail throughout that session. Problems ensue when a user walks away from a terminal and another user takes over, or when an authorised user deliberately hands over to an unauthorised one. A variety of attempts have been made to counter this using \emph{continuous authentication}. Such approaches may prove of significant use where cobots are concerned. Hass, Ulz and Steger \cite{haas2017secured} approach this problem with expiring passwords and smart cards equipped with fingerprint readers, so only authorised users are authenticated to use the robot.

\subsubsection{Intrusion Detection Systems~(IDS).}
Intrusion Detection Systems are effective counter-measures for detecting attacks or improper use of systems by detecting any anomalies or inappropriate use of the host machines or networks. The concept of IDSs was first introduced by Anderson in 1980 \cite{anderson1980}. There are three main intrusion detection approaches: signature-based, anomaly-based, and specification-based \cite{lee1999,sekar2002specification,liao2013intrusion}. 

\emph{Signature-based systems} recognise basic pre-packaged patterns of misbeha\-viour. For example, three consecutive failed attempts to log into a cobot management system might be regarded as a `signature' of a potential attack. Another example includes the use of specific payloads in service requests recognised by the presence of specific bit strings in code. The bit string would form a signature of malfeasance. 

An \emph{anomaly-based approach} typically seeks to profile `normal’
behaviour in some way and measure how closely current behaviour is to
that previously ascertained profile. Often, the underlying measure is a
statistical one, and  an anomaly is raised when current behaviour
veers outside the historically established profile of normal
behaviour. Some attributes lend themselves to monitoring in this way,
for example many events incurred, such as the number of page faults or cache-misses. Often an anomaly-based system is required to classify current behaviour as `normal' or `anomalous'. Thus, it is no surprise that a wide variety of pattern classification approaches can, and have, been brought to bear on intrusion detection problems. A major claim for anomaly-based approaches is that they have good potential for detecting previously unseen attacks. The degree to which this is true is controversial; strictly speaking, they will flag as anomalous monitored behaviours that are not reasonably close to historic trends. However, some malicious attacks may have behaviours that are plausibly consistent with normal behaviours. If the symptoms of an attack are similar to the symptoms of another anomalous attack then the new attack will also be flagged as anomalous. Furthermore, most behaviours are unique if you look into the detail. Thus, such approaches require that, at some level of detail, attacks look different to `normal' behaviour. This may or may not be true.  

\emph{Specification-based intrusion detection} \cite{sekar2002specification}, being a niche application, assumes
that the behaviour follows a specified protocol. An intrusion is any deviation from this protocol. A variety of attacks on communication or application
protocols can be detected this way.  For example, a malicious
application running on a cobot may attempt to make use of currently unused fields in a communication protocol to leak information. Such leaking will be detectable.

\label{intrusion2}
An example of the application of intrusion detection techniques to a robotic system is given in~\cite{jones2017using}, where the authors implement an IDS using deep learning models. There are two components in the system proposed: one signature-based and one anomaly-based. The  signature-based component is intended to detect misuse by identifying known malicious activity signatures, whilst the anomaly-based component detects behavioural anomalies by analysing deviations from the expected behaviour.  The approach has two significant challenges: processing time and detection accuracy. In short, the IDS must be very fast with a low false positive rate. This is acceptable when using the signature-based model, but  the anomaly detection component requires more time and is more prone to falsely indicating an attack. 

In another instance, Fagiolini, Dini and Bicchi \cite{fagiolini2014distributed} propose an intrusion detection system that detects misbehaviour in systems where robots interact on the premise of event-based rules. They refer to misbehaving robots as intruders that may exhibit uncooperative behaviour due to spontaneous failure or malicious reprogramming. The proposed IDS is applied in an industrial scenario where a number of automated forklifts move within an environment that can be represented as a matrix of cells and macro-cells.

\subsection{Application to Cobots}
\label{sec:example:sec}

To the best of our knowledge, there would appear to be no published research specifically targeting security for cobots. However, the authors are currently undertaking work into the application of security concepts described above. In the next section we identify some of these and explain why the application to cobotic security may require some sophistication.

\subsubsection{Security Policy.}

Since the rise of cobots is quite recent we should not be surprised to
find no security policy work specifically targeting cobots. This area,
however, is an intriguing and subtle one. A security policy will
generally have to maintain some notion of privacy, as personal data
may be stored in the system~(e.g. a staff capabilities database). But
the area is much more subtle than is usual for security policy.  Issues of who should get access to what and
under what circumstances, may have significant effect on co-workers. What is a reasonable policy on
access when there is contention between co-workers? What is suitable
'etiquette' in such cases? 
Some data will be less privacy-sensitive and safety-relevant and
the motivations for security policy elements will be different. 
A security policy and its implementation may be able to enforce elements
of safety practice. For example, overly long shifts that break working time directives can be readily policed via security policy. Requirements for suitable skills or training can be enforced too. Ultimately, a great deal of reasonable co-working constraints boil down to access control of one form or another and such control can be specified via policy.

\subsubsection{Authentication.}

Our current work is carried out in collaboration with industrial users
of cobots. Workers generally operate a single robot cell and  work on
a continual basis, punctuated by scheduled breaks. It seems highly
unlikely that the robots in these cells can be faked or replaced in
some way without notice~(though we accept that software could be
maliciously updated). Consequently, it is only one-way user-to-robot
authentication that needs to be considered. However, this
authentication needs to be persistent, to account for changes in
staffing. Some notion of initial authentication by traditional means,
for example a standard password or token based approach, can be supplemented by an appropriate continuous authentication method. We are currently investigating the use of physical interaction properties~(i.e. forces applied to the robot by the co-worker) and other biometrics~(accelerometry applied to co-worker hands, eye movements and ECG monitoring) for user authentication. 

The resources and capabilities made available by a cobot may
vary across co-workers.  Part of the appeal of cobots is, however,
that they can be readily reprogrammed to do a variety of tasks. As
co-workers may vary in their skills, experience and qualifications, we
might quite legitimately allow certain tasks to be done only by
co-workers with specific skills and training. A suitably configured
system with access to the capabilities of specific users can easily
enforce such policies. Thus, issues of cobot policy compliance can
readily be handled via suitable access control, provided there is
credible user authentication.

\subsection{Challenges for Cobot Security}

Many aspects of cybersecurity analysis and system development remain unaltered. But our work so far reveals that cobots also present either specific challenges in their own right, or else require considerable subtlety of thought to engage with cobot-specifics. Below we highlight areas we believe are in need of security focus:

\emph{Developing security policies} and related elements that fully take into account the human, including both single co-worker cases and multi-co-worker cases. Policies must be sympathetic of user needs, well-being, and diversity issues. Crafting a policy for a mobile cobot working in different company domains is as much an exercise in occupational psychology as it is an issue in cyber-security.  Particularly, in an industrial work setting, the co-working modus operandi and co-worker well-being must be considered. There are significant possibilities of couching a variety of working practice constraints in terms of security policy.

\emph{Development of practical templates for authentication
  requirements}. Authentication needs will vary a great deal between
deployment domains. Thus, a single worker collaborating with an
industrial robot on a welding task might lend itself to initial
password logon followed by continuous authentication by a variety of
means~(e.g. biometrics or token). Manufacturing with several on-site suppliers and changing external staffing is markedly
different, perhaps requiring on-going speech recognition for
continuous authentication. Guidance will need to be developed about
the pros and cons of authentication approaches in specific cobot
environments. These will include 
guidance regarding specific approaches in harsh electromagnetic
environments.  As an example of the latter, one of our industrial
collaborators uses cobots to perform elements of arc-welding; the
arc-ing could affect any authentication scheme based around wireless
communications~(including, e.g.\ distance bounding protocols).

\emph{Development of a plausible strategy for \textbf{cobot
    forensics}}, which as far as we are aware, has received no
attention in the literature. Digital forensics is largely concerned
with the generation of credible evidence as to who did what, where and
when (and also why). This is important not just for cyber-security
reasons, for example to investigate security breaches, but also for
health and safety reasons, where reconstructing an accident from event
logs may be required.

\emph{Developing a credible IDS} for use in cobotic
environments. Again, we know of no work in this area. It is clear
that domain specifics will need to be handled. Mobile cobots will
present challenges over and above those for fixed cobots.

\section{Cobot Co-Assurance} 
\label{sec:safesecure}

In previous sections, we have discussed how to reduce safety and security risks in cobots, and the challenges that arise. However, reducing risk within the individual domains alone is insufficient to claim the acceptability of overall risk. %
There are multiple factors that can affect co-assurance and the confidence in the reduction of overall risk. These factors can be divided into two categories:
\begin{description}
\item[Socio-technical Factors] -- these are concerned with the processes, technology, structure and people required for co-engineering and the \textit{co-assurance process}. Such factors play an important role, e.g. in complex information systems engineering~\cite{bostrom1977mis}, and so co-assuring socio-technical factors encompasses a large scope that includes organisational, regulatory, and ethical structures, management practices, competence, information management tools, \textit{etc.}
\item[Technical Factors] -- these are primarily concerned with the causal relationships between risk conditions and artefacts of the \textit{engineered system}, \textit{e.g.} accidents, hazards, attacks, threats, safety and security requirements. 
\end{description}

Below, we consider these factors for cobots, and discuss the challenges that arise with respect to the interactions between safety and security risks.

\subsection{Socio-technical Challenges} %

The many socio-technical challenges include: 

\emph{Ethics.} Interaction with co-workers is at the core of cobot use and ethics issues will inevitably arise.  For example, ethical issues related to operator monitoring within the cobot will have to be resolved, for example by getting informed consent from co-workers.
    
\emph{Risk prioritisation.} The principled handling of trade-offs across competing risks in cobots (safety, security, economic etc.) requires care in both  `normal' and exceptional circumstances. For example, when a security problem has been published with potential safety implications but a vendor patch is not yet available (possibly a common occurrence given the heavy use of  commercially available components in cobots), what should be done immediately? Different actions may have different economic consequences and organisations are rarely equipped to judge such situations rapidly.

\emph{Risk representation and communication.} To facilitate better decision making, the \emph{profile} of various risks in particular circumstances needs to be made apparent, e.g. using structured risk analysis~(\Cref{sec:str-rsk-ana}). This may encompass risk attributes outside the usual safety and security~(or further dependability).   

\emph{Coordinating and integrating risk analyses.} The system design
may be modified, as the design and analyses progress, to incorporate
necessary security and safety mechanisms. Since neither security nor
safety are preserved under refinement (i.e. moving to a more concrete
representation) the \emph{specific way} a safety or security measure
is implemented matters and so safety and security synchronisation will
be required at what are believed to be at feasibly stable points in
the development.  \Cref{tab:approachesTable} on page~\pageref{tab:approachesTable} summarises several candidate approaches
to be adopted to prioritise, align, or integrate safety/security analyses.

\emph{Ensuring quality and availability of third party evidence.} Cobots will usually integrate components from several vendors. Those components  may draw on further vendors' products. Ensuring the quality or availability of relevant documentation throughout the supply chain to support safety and security assurance arguments, both individually and together, due to interdependencies, will often prove challenging. A vendor patch to a vulnerability may, for example, be available only as an executable.
    
\emph{Handling dynamic threat and impact landscapes.} A cobot's threat landscape may vary greatly over time, e.g. when new vulnerabilities in critical components emerge, and this will likely affect safety. Rapidly assessing the safety impact of security flaws~(and vice versa) is not a mainstream activity for manufacturing organisations and the skills required may well be lacking.

\emph{Resourcing safety and security interactions.} There are many interactions between safety and security in cobots. Resourcing the investigation of such aspects will present significant challenges in the workplace.
    
\emph{Increasing automation of assurance.}  Cobot assurance is complex
and dynamic, requiring repeated analyses and risk judgements to be
frequently revisited. Increasing automation of assurance will be
essential to ensure feasibility of any sound process.
\Cref{tab:approachesTable}~(rightmost
column) on page~\pageref{tab:approachesTable} lists approaches that, used individually or in combination,
can be a basis for continuous alignment of safety and security assurance
of cobots in manufacturing.

\subsection{Technical Challenges}

Some technical challenges in cobot co-assurance are given below:

\emph{Deadly conflicts.} Cobots may be used in dangerous environments with human presence. Protecting against dangerous, and possibly deadly, conflicts is a challenge. For example, ensuring access control policy does not cause problems, i.e. by prohibiting essential data access for maintaining safety. 
    
\emph{Consolidated audit policy.} Logging events are an important, if
somewhat prosaic, component of cobots. Many events \emph{could}
be logged but the potential volume is huge. We need to determine for
what  purposes we would want logged information  and seek to craft a
fitting policy. For security purposes, we might wish to access logs
for determining situational awareness, for identifying specific  behaviours, or for contributing towards evidence in a criminal court case. For safety purposes, we might wish to reconstruct an accident and its causes. We must be aware of the possibility of logging  causing  problems~(e.g. affecting  real-time performance of critical  services or, even worse, causing deadlocks). Of course, the logs themselves must be protected from attack, as a successful security attack might destroy forensic log evidence. The topic of \emph{attribution}  is difficult in almost any system, however the complexity of cobots and their wider connectedness will make this aspect even hard for cobots. 
    
\emph{Unifying anomaly detection.} There are challenges regarding the \emph{efficient} use of data to inform anomaly detection for safety and security as the data that can inform safety and security decisions may overlap. For example, the presence of sophisticated malware might reveal itself in degradation trends of operational performance. Thus, traditional health monitoring for safety and predictive maintenance reasons might  highlight the presence of malware~(one of Stuxnet's remarkable attack goals was to cause centrifuges to wear out more quickly \cite{falliere2011w32}).
    
\emph{Reacting to compromise.} What should be done when some elements of a cobot are perceived to be compromised in some way~(e.g. via physical component failure or the presence of malware)? Having detected something is potentially `wrong', management may wish to take appropriate action. This requires significant situational awareness~(which may be difficult in cobots) and we need to be very careful  to ensure the cure~(response) is not worse than the problem, i.e. does not itself cause unpalatable security or safety issues. Some degree of degradation may be inevitable.
    
\emph{Ensuring resilience to direct access.} Cobots will  come into close proximity with humans providing an opportunity for  interference or damage, both physical and digital. We must find useful and preferably common mechanisms to either resist attack or else make it apparent when it has occurred. 
    
\emph{Coping under failure.} The failure modes of a cobot and
their consequences must be thoroughly understood. A degraded system
may be compromisable in ways a fully operational system is not.
Co-assurance approaches from process automation such as,
e.g.\ FACT~\cite{Sabaliauskaite2015-AligningCyberPhysical}, could cross-fertilise
cobot-specific co-assurance procedures.
    
\emph{Adversarial attacks.}  Systems whose activities are underpinned by machine learning or other AI\footnote{Cobots, as with other manufacturing technologies,  will include increasingly complex artificial intelligence software.} may be susceptible to \emph{adversarial attack}. This is where the robot is fooled into malfunctioning in some way because it is presented with specially crafted input. This is a common worry from both a security and safety perspective. 
    
\emph{Testing of cobots.} Current robot testing approaches will need to be adapted to the new requirements of cobot settings.  Moreover, the development of more rigorous testing approaches for cobots is essential.
    
\emph{Software and system update.} Cobot and wider system update is a common concern and we need to ensure updates are made with appropriate authority, integrity, and that the updates do not cause unpalatable harm. From a security perspective, patches are a commonplace, however the safety implications of such changes are rarely considered. Other domains understand the importance of maintaining integrity of updates~(e.g.\ updating of remote satellites or automotive software in the field).  In some cases updates may need to be `hot', i.e.\ at run-time. Again, a compromised update process will play havoc with safety and security. Furthermore, the effects of a host of run-time attacks are likely to have safety implications~(e.g.\ attacks on a database via SQL injection, or similar). 

\subsection{Conclusion}
\label{sec:conc}

With this chapter, we summarise general approaches to safety and
security assurance for cobots. Further detail is provided in the form
of risk-based assurance approaches for both safety and security of
cobots~(\Cref{sec:generalApproaches}), illustrated by an example~(\Cref{sec:example:sec,sec:appl-cobots-proj}).  Finally, we describe the
challenges that arise when attempting to assure and align the two
quality attributes for a cobot. These challenges are not
limited to technical factors about how and when to relate risk
conditions between safety and security, but include significant
socio-technical factors that influence the co-assurance process
itself.
These challenges form the crux of the contribution of this
work. Although no claims are made with regards to completeness, we
believe that the challenges identified in \Cref{sec:safesecure}
form the basis of a preliminary research roadmap, and that addressing them
is essential for the safe and secure deployment of cobots in an
industrial setting.

\paragraph{Acknowledgements.}

Contribution to this chapter was made through the Assuring Autonomy International Programme (AAIP) Project - CSI:Cobot. The AAIP is funded by Lloyds Register Foundation.

\bibliographystyle{./admin/splncs04}
\bibliography{}

\begin{thebibliography}{10}
\providecommand{\url}[1]{\texttt{#1}}
\providecommand{\urlprefix}{URL }
\providecommand{\doi}[1]{https://doi.org/#1}

\bibitem{abdo2018safety}
Abdo, H., Kaouk, M., Flaus, J.M., Masse, F.: A safety/security risk analysis
  approach of industrial control systems: A cyber bowtie--combining new version
  of attack tree with bowtie analysis. Computers \& Security  \textbf{72},
  175--195 (2018)

\bibitem{Alami2006-Safedependablephysical}
Alami, R., Albu-Schaeffer, A., Bicchi, A., Bischoff, R., Chatila, R., Luca,
  A.D., Santis, A.D., Giralt, G., Guiochet, J., Hirzinger, G., Ingrand, F.,
  Lippiello, V., Mattone, R., Powell, D., Sen, S., Siciliano, B., Tonietti, G.,
  Villani, L.: Safe and dependable physical human-robot interaction in
  anthropic domains: State of the art and challenges. In: 2006 {IEEE}/{RSJ}
  International Conference on Intelligent Robots and Systems. {IEEE} (2006).
  \doi{10.1109/iros.2006.6936985}

\bibitem{anderson1980}
Anderson, J.P.: Computer security threat monitoring and surveillance. Technical
  Report, James P. Anderson Company  (1980)

\bibitem{arkin2005software}
Arkin, B., Stender, S., McGraw, G.: Software penetration testing. IEEE Security
  \& Privacy  \textbf{3}(1),  84--87 (2005)

\bibitem{bostrom1977mis}
Bostrom, R.P., Heinen, J.S.: Mis problems and failures: A socio-technical
  perspective. part i: The causes. MIS quarterly pp. 17--32 (1977)

\bibitem{bouti1994state}
Bouti, A., Kadi, D.A.: A state-of-the-art review of fmea/fmeca. International
  Journal of reliability, quality and safety engineering  \textbf{1}(04),
  515--543 (1994)

\bibitem{cook2016measuring}
Cook, A., Smith, R., Maglaras, L., Janicke, H.: Measuring the risk of cyber
  attack in industrial control systems. In: 4th International Symposium for ICS
  \& SCADA Cyber Security Research 2016 (ICS-CSR). BCS eWiC (2016).
  \doi{10.14236/ewic/ics2016.12}

\bibitem{despotou2009addressing}
Despotou, G., Alexander, R., Kelly, T.: Addressing challenges of hazard
  analysis in systems of systems. In: 2009 3rd Annual IEEE Systems Conference.
  pp. 167--172. IEEE (2009)

\bibitem{dieber2017security}
Dieber, B., Breiling, B., Taurer, S., Kacianka, S., Rass, S., Schartner, P.:
  Security for the robot operating system. Robotics and Autonomous Systems
  \textbf{98},  192--203 (2017)

\bibitem{dunjo2010hazard}
Dunj{\'o}, J., Fthenakis, V., V{\'\i}lchez, J.A., Arnaldos, J.: Hazard and
  operability ({HAZOP}) analysis. a literature review. Journal of hazardous
  materials  \textbf{173}(1-3),  19--32 (2010)

\bibitem{eames1999integration}
Eames, D.P., Moffett, J.: The integration of safety and security requirements.
  In: International Conference on Computer Safety, Reliability, and Security.
  pp. 468--480. Springer (1999)

\bibitem{fagiolini2014distributed}
Fagiolini, A., Dini, G., Bicchi, A.: Distributed intrusion detection for the
  security of industrial cooperative robotic systems. IFAC Proceedings Volumes
  \textbf{47}(3),  7610--7615 (2014)

\bibitem{falliere2011w32}
Falliere, N., Murchu, L.O., Chien, E.: W32. stuxnet dossier. White paper,
  Symantec Corp., Security Response  \textbf{5}(6), ~29 (2011)

\bibitem{friedberg2017stpa}
Friedberg, I., McLaughlin, K., Smith, P., Laverty, D., Sezer, S.:
  {STPA-SafeSec}: Safety and security analysis for cyber-physical systems.
  Journal of Information Security and Applications  \textbf{34},  183--196
  (2017)

\bibitem{gilchrist1993modelling}
Gilchrist, W.: Modelling failure modes and effects analysis. International
  Journal of Quality \& Reliability Management  (1993)

\bibitem{haas2017secured}
Haas, S., Ulz, T., Steger, C.: Secured offline authentication on industrial
  mobile robots using biometric data. In: Robot World Cup. pp. 143--155.
  Springer (2017)

\bibitem{handbook1981nureg}
Handbook, F.T., Roberts, N., Vesely, W., Haasl, D., Goldberg, F.: Nureg-0492.
  US Nuclear Regulatory Commission  (1981)

\bibitem{Hayes2013-ChallengesSharedEnvironment}
Hayes, B., Scassellati, B.: Challenges in shared-environment human-robot
  collaboration. In: Proceedings of the Collaborative Manipulation Workshop at
  HRI (2013)

\bibitem{he2012generic}
He, Y., Johnson, C.: Generic security cases for information system security in
  healthcare systems. In: 7th {IET} International Conference on System Safety,
  incorporating the Cyber Security Conference 2012. IET (2012).
  \doi{10.1049/cp.2012.1507}

\bibitem{hinojosa2018advanced}
Hinojosa, C., Potau, X.: Advanced industrial robotics: Taking human-robot
  collaboration to the next level. Eurofound \& European Commission, Brussels,
  Belgium (2018)

\bibitem{IFR2018demistifying}
{IFR}: Demystifying Collaborative Industrial Robots. International Federation
  of Robotics, Frankfurt, Germany (2018)

\bibitem{ISO10218}
{ISO 10218}: Robots and robotic devices -- safety requirements for industrial
  robots. Standard, Robotic Industries Association (RIA) (2011),
  \url{https://www.iso.org/standard/51330.html}

\bibitem{ISO15066}
{ISO/TS 15066}: Robots and robotic devices -- collaborative robots. Standard,
  Robotic Industries Association (RIA) (2016),
  \url{https://www.iso.org/standard/62996.html}

\bibitem{johnson2019assurance}
Johnson, N., Kelly, T.: An assurance framework for independent co-assurance of
  safety and security. In: Muniak, C. (ed.) Journal of System Safety.
  International System Safety Society (January 2019), presented at: the 36th
  International System Safety Conference (ISSC). Arizona, USA: August 2018

\bibitem{johnson2019through}
Johnson, N., Kelly, T.: Devil's in the detail: Through-life safety and security
  co-assurance using ssaf. In: International Conference on Computer Safety,
  Reliability, and Security. Springer (2019)

\bibitem{jones2017using}
Jones, A., Straub, J.: Using deep learning to detect network intrusions and
  malware in autonomous robots. In: Cyber Sensing 2017. vol. 10185, p. 1018505.
  International Society for Optics and Photonics (2017)

\bibitem{Jones1986-StudySafetyProduction}
Jones, R.H.: A Study of Safety and Production Problems and Safety Strategies
  Associated with Industrial Robot Systems. Ph.D. thesis, Imperial College
  (1986)

\bibitem{Kaiser2018-SafetyRelatedRisks}
Kaiser, L., Schlotzhauer, A., Brandstötter, M.: Safety-related risks and
  opportunities of key design-aspects for industrial human-robot collaboration.
  In: Lecture Notes in Computer Science, pp. 95--104. Springer International
  Publishing (2018). \doi{10.1007/978-3-319-99582-3_11}

\bibitem{kazman1998architecture}
Kazman, R., Klein, M., Barbacci, M., Longstaff, T., Lipson, H., Carriere, J.:
  {The Architecture Tradeoff Analysis Method}. In: Proceedings. Fourth IEEE
  International Conference on Engineering of Complex Computer Systems (Cat. No.
  98EX193). pp. 68--78. IEEE (1998)

\bibitem{kelly1999arguing}
Kelly, T.P.: Arguing safety: a systematic approach to managing safety cases.
  Ph.D. thesis, University of York York, UK (1999)

\bibitem{knowles2015survey}
Knowles, W., Prince, D., Hutchison, D., Disso, J.F.P., Jones, K.: A survey of
  cyber security management in industrial control systems. International
  journal of critical infrastructure protection  \textbf{9},  52--80 (2015)

\bibitem{kohnfelder1999threats}
Kohnfelder, L., Garg, P.: The threats to our products. Microsoft Interface,
  Microsoft Corporation  \textbf{33} (1999)

\bibitem{kordy2010foundations}
Kordy, B., Mauw, S., Radomirovi{\'c}, S., Schweitzer, P.: Foundations of
  attack--defense trees. In: International Workshop on Formal Aspects in
  Security and Trust. pp. 80--95. Springer (2010)

\bibitem{lee1999}
Lee, W., Stolfo, S.J., Mok, K.W.: A data mining framework for building
  intrusion detection models. In: Proceedings of the 1999 IEEE Symposium on
  Security and Privacy (Cat. No. 99CB36344). pp. 120--132. IEEE (1999)

\bibitem{leveson2003new}
Leveson, N.G.: A new approach to hazard analysis for complex systems. In:
  International Conference of the System Safety Society (2003)

\bibitem{Leveson2012-EngineeringSaferWorld}
Leveson, N.G.: Engineering a Safer World: Systems Thinking Applied to Safety.
  Engineering Systems, MIT Press (2012). \doi{10.7551/mitpress/8179.001.0001}

\bibitem{liao2013intrusion}
Liao, H.J., Lin, C.H.R., Lin, Y.C., Tung, K.Y.: Intrusion detection system: A
  comprehensive review. Journal of Network and Computer Applications
  \textbf{36}(1),  16--24 (2013)

\bibitem{lichte2018use}
Lichte, D., Wolf, K.: Use case-based consideration of safety and security in
  cyber physical production systems applied to a collaborative robot system.
  In: Safety and Reliability--Safe Societies in a Changing World, pp.
  1395--1401. CRC Press (2018)

\bibitem{maggi2017rogue}
Maggi, F., Quarta, D., Pogliani, M., Polino, M., Zanchettin, A.M., Zanero, S.:
  Rogue robots: Testing the limits of an industrial robot’s security. Trend
  Micro, Politecnico di Milano, Tech. Rep  (2017)

\bibitem{martin2018quantitative}
Mart{\'\i}n, F., Soriano, E., Ca{\~n}as, J.M.: Quantitative analysis of
  security in distributed robotic frameworks. Robotics and Autonomous Systems
  \textbf{100},  95--107 (2018)

\bibitem{matheson2019human}
Matheson, E., Minto, R., Zampieri, E.G., Faccio, M., Rosati, G.: Human--robot
  collaboration in manufacturing applications: A review. Robotics
  \textbf{8}(4), ~100 (2019)

\bibitem{Nicolaisen1985-OccupationalSafetyIndustrial}
Nicolaisen, P.: Occupational safety and industrial robots. In: Bonney, Yong
  (eds.) Robot Safety, pp. 33--48. IFS (Publications) Ltd. (1985).
  \doi{10.1007/978-3-662-02440-9_9}

\bibitem{papadopoulos1999hierarchically}
Papadopoulos, Y., McDermid, J.A.: Hierarchically performed hazard origin and
  propagation studies. In: International Conference on Computer Safety,
  Reliability, and Security. pp. 139--152. Springer (1999)

\bibitem{pawar2016manufacturing}
Pawar, V.M., Law, J., Maple, C.: Manufacturing robotics: The next robotic
  industrial revolution. UK-RAS White papers, UK-RAS Network (2016).
  \doi{10.31256/wp2016.1}

\bibitem{pietre2013cross}
Pi{\`e}tre-Cambac{\'e}d{\`e}s, L., Bouissou, M.: Cross-fertilization between
  safety and security engineering. Reliability Engineering \& System Safety
  \textbf{110},  110--126 (2013)

\bibitem{quigley2009ros}
Quigley, M., Conley, K., Gerkey, B., Faust, J., Foote, T., Leibs, J., Wheeler,
  R., Ng, A.Y.: {ROS}: an open-source robot operating system. In: ICRA workshop
  on open source software. vol.~3.2, p.~5. Kobe, Japan (2009)

\bibitem{Sabaliauskaite2015-AligningCyberPhysical}
Sabaliauskaite, G., Mathur, A.P.: Aligning cyber-physical system safety and
  security. In: Cardin, M.A., Krob, D., Lui, P.C., Tan, Y.H., Wood, K. (eds.)
  Complex Systems Design \& Management Asia, pp. 41--53. Springer (2015).
  \doi{10.1007/978-3-319-12544-2_4}

\bibitem{schmittner2014fmvea}
Schmittner, C., Ma, Z., Smith, P.: Fmvea for safety and security analysis of
  intelligent and cooperative vehicles. In: International Conference on
  Computer Safety, Reliability, and Security. pp. 282--288. Springer (2014)

\bibitem{schneier1999attack}
Schneier, B.: Attack trees. Dr. Dobb’s journal  \textbf{24}(12),  21--29
  (1999)

\bibitem{sekar2002specification}
Sekar, R., Gupta, A., Frullo, J., Shanbhag, T., Tiwari, A., Yang, H., Zhou, S.:
  Specification-based anomaly detection: a new approach for detecting network
  intrusions. In: Proceedings of the 9th ACM conference on Computer and
  communications security. pp. 265--274 (2002)

\bibitem{shostack2014threat}
Shostack, A.: Threat modeling: Designing for security. John Wiley \& Sons
  (2014)

\bibitem{ros2_Robotic_systems_threat_model}
{Special Interest Group on Next-Generation ROS}: Ros 2 robotic systems threat
  model, \url{https://design.ros2.org/articles/ros2\_threat\_model.html}

\bibitem{srivatanakul2004effective}
Srivatanakul, T., Clark, J.A., Polack, F.: Effective security requirements
  analysis: {HazOp} and use cases. In: International Conference on Information
  Security. pp. 416--427. Springer (2004)

\bibitem{Sugimoto1977-SafetyEngineeringIndustrial}
Sugimoto, N.: Safety engineering on industrial robots and their draft standards
  for safety requirements. In: Proceedings of the 7th International Symposium
  on Industrial Robots. pp. 461--470 (1977)

\bibitem{tang2019human}
Tang, G., Webb, P.: Human-robot shared workspace in aerospace factories. In:
  Human-robot interaction: safety, standardization, and benchmarking, pp.
  71--80. CRC Press (2019)

\bibitem{tantawi2019advances}
Tantawi, K.H., Sokolov, A., Tantawi, O.: Advances in industrial robotics: From
  industry 3.0 automation to industry 4.0 collaboration. In: 2019 4th
  Technology Innovation Management and Engineering Science International
  Conference (TIMES-iCON). pp.~1--4. IEEE (2019)

\bibitem{vanderborght2019unlocking}
Vanderborght, B.: Unlocking the Potential of Industrial Human–Robot
  Collaboration. Publications Office of the European Union, Brussels, Belgium
  (2019)

\bibitem{vilches_2019}
Vilches, V.M.: Threat modeling a ros 2 robot (Oct 2019),
  \url{https://news.aliasrobotics.com/threat-modeling-a-ros-2-robot/}

\bibitem{winther2001security}
Winther, R., Johnsen, O.A., Gran, B.A.: Security assessments of safety critical
  systems using {HAZOP}s. In: International Conference on Computer Safety,
  Reliability, and Security. pp. 14--24. Springer (2001)

\bibitem{young2014integrated}
Young, W., Leveson, N.G.: An integrated approach to safety and security based
  on systems theory. Commun. ACM  \textbf{57}(2),  31--35 (2014)

\bibitem{yu2017bow}
Yu, M., Venkidasalapathy, J., Shen, Y., Quddus, N., Mannan, S.M., et~al.:
  Bow-tie analysis of underwater robots in offshore oil and gas operations. In:
  Offshore Technology Conference. Offshore Technology Conference (2017)

\bibitem{zong2019policy}
Zong, Y., Guo, Y., Chen, X.: Policy-based access control for robotic
  applications. In: 2019 IEEE International Conference on Service-Oriented
  System Engineering (SOSE). pp. 368--3685. IEEE (2019)

\end{thebibliography}

\end{document}